\documentclass{article}

\usepackage{silence}
\usepackage{amsmath}
\usepackage{amssymb}
\usepackage{longtable}
\usepackage{booktabs}
\usepackage{bm}
\usepackage{stmaryrd}
\usepackage{stfloats}
\usepackage{tikz, pgfplots}
\usetikzlibrary{shapes,arrows}
\usepackage{amsthm}
\usepackage{textcomp}
\pgfplotsset{compat=1.14}
\usepackage{enumitem}
\setlist{nosep,leftmargin=1em}

\usepackage{microtype}
\usepackage{graphicx}
\usepackage{subfigure}
\usepackage{booktabs} 

\usepackage{hyperref}

\DeclareMathOperator*{\argmax}{arg\,max}

\newtheorem{definition}{Definition}
\newtheorem{prop}{Proposition}

\usepackage[accepted]{icml2020}

\icmltitlerunning{Unbiased Risk Estimators Can Mislead: A Case Study of Learning with Complementary Labels}

\usepackage{soul}

\begin{document}

\twocolumn[

\icmltitle{Unbiased Risk Estimators Can Mislead:\\
    A Case Study of Learning with Complementary Labels}



\icmlsetsymbol{intern}{*}

\begin{icmlauthorlist}
\icmlauthor{Yu-Ting Chou}{ntu,intern}
\icmlauthor{Gang Niu}{riken}
\icmlauthor{Hsuan-Tien Lin}{ntu}
\icmlauthor{Masashi Sugiyama}{riken,utokyo}
\end{icmlauthorlist}

\icmlaffiliation{ntu}{National Taiwan University}
\icmlaffiliation{riken}{RIKEN}
\icmlaffiliation{utokyo}{The University of Tokyo}

\icmlcorrespondingauthor{Yu-Ting Chou}{r07922042@csie.ntu.edu.tw}

\icmlkeywords{Machine Learning, ICML}

\vskip 0.3in
]



\printAffiliationsAndNotice{*Work done during an internship at RIKEN.}  

\begin{abstract}
In weakly supervised learning, \emph{unbiased risk estimator}~(URE) is a powerful tool for training classifiers when training and test data are drawn from different distributions.
Nevertheless, UREs lead to overfitting in many problem settings when the models are complex like deep networks.
In this paper, we investigate reasons for such overfitting by studying a weakly supervised problem called \emph{learning with complementary labels}.
We argue the quality of \emph{gradient estimation} matters more in risk minimization.
Theoretically, we show that a URE gives an \emph{unbiased gradient estimator}~(UGE).
Practically, however, UGEs may suffer from huge variance, which causes empirical gradients to be usually far away from true gradients during minimization.
To this end, we propose a novel \emph{surrogate complementary loss}~(SCL) framework that trades zero bias with reduced variance and makes empirical gradients more aligned with true gradients in the direction.
Thanks to this characteristic, SCL successfully mitigates the overfitting issue and improves URE-based methods.
\end{abstract}

\section{Introduction}
In \emph{weakly supervised learning}~(WSL), learning algorithms have to train classifiers under incomplete, inexact or inaccurate supervision \cite{zhou2017brief},
including but not limited to
semi-supervised learning \cite{chapelle2009semi},
partial labels \cite{jin2002partial},
noisy labels \cite{natarajan2013learning, patrini2017making, han2018masking, han2018coteaching, yu2019coplus, xia2019anchor},
complementary labels \cite{ishida2017learning, yu2018learning, ishida2019complementary, xu2019generative, feng2019learning},
where the label distribution changes, and
positive-unlabeled data \cite{elkan2008learning, du2014analysis, du2015convex, niu2016theoretical, pmlr-v70-sakai17a, sakai2018semi},
unlabeled-unlabeled data \cite{lu2018minimal, lu2019mitigating},
and other similar settings \cite{bao2018classification,ishida2018binary,hsieh2019classification}, 
where the data distribution changes.
Among WSL methods, \emph{unbiased risk estimator}~(URE) is a powerful tool:
it evaluates the \emph{classification risk} from training data drawn from a distribution different from the test one, and thus \emph{empirical risk minimization} \cite{vapnik1992principles} is possible.
The success of URE is due to two orthogonal demands in WSL for handling big data and complex data:
URE poses \emph{unconstrained optimizations} so that it can handle very big data by \emph{stochastic optimizers};
URE is \emph{model-independent} so that it can handle complex data where the model is chosen according to the data (e.g., image, text, or speech).

An important motivation of employing URE in WSL is that URE enables \emph{estimation error bounds} to guarantee \emph{statistical consistency}.
However, the consistency in the \emph{asymptotic cases} is not very meaningful in the \emph{finite-sample cases} especially in deep learning \cite{zhang2016understanding, nagarajan2019uniform}.
Despite its popularity and nice properties, URE in \citet{du2015convex}, \citet{ishida2017learning} or \citet{lu2018minimal} has inferior test performance to recent biased methods in \citet{kiryo2017positive}, \citet{ishida2019complementary} and \citet{lu2019mitigating}.
When complex models like deep networks are chosen as the classifiers, UREs suffer from severe \emph{negative empirical risks} during training, which is a sign of overfitting.
Even though the overfitting issue can be relatively mitigated by keeping UREs non-negative, the mechanism behind how UREs cause overfitting is still unknown.
Thus, instead of a theoretical motivation, this paper has a practical motivation and focuses on understanding how UREs cause overfitting and how to avoid such overfitting in algorithm design.

\emph{Learning with complementary labels} \cite{ishida2017learning} is a WSL problem of multi-class classification where classifiers are trained from data with complementary labels~(CL).
A CL specifies a class that an instance \emph{does not belong to}, but the trained classifier should still predict the correct labels.
Although CLs are less informative than ordinary labels, they provide an alternative when ordinary labels are inaccessible or costly to acquire.
In this paper, we choose learning with CLs to study the overfitting issue of UREs, as it combines several practical advantages:
first, CLs are easy to \emph{generate} compared with partial labels and noisy labels;
second, negative empirical risks are easy to \emph{occur};
and third, it is easy to experimentally \emph{analyze} the \emph{bias} and \emph{variance} of empirical gradients.
With the help of such a case study, we can gain a deep insight of UREs and lay the foundation for further studies of UREs in other WSL problem settings.

Our contributions can be summarized in two folds.
First of all, we conduct a series of analyses to investigate reasons for the overfitting issue.
We show that due to the linearity of the differential operator, any URE must give an \emph{unbiased gradient estimator}~(UGE);
however, UGE is not necessarily good at gradient estimation though it is unbiased.
During training, only a single fixed CL could be acquired for each instance, which causes empirical gradients given by a UGE to be usually far away from true gradients.
This illustrates the difference between \emph{validation} and \emph{training}:
\begin{itemize}
    \item In validation, the classifier is fixed and the data is repeatedly sampled, and then UGE is good at gradient estimation (which can be theoretically guaranteed by \emph{concentration inequalities}).
    \item In training, the data are fixed and based on these data the classifier is iteratively updated, and then UGE might be really bad at gradient estimation.
    \item Theoretically speaking, good validation can imply good training if the model is simple, while good validation may still result in poor training if the model is complex \cite{zhang2016understanding, nagarajan2019uniform}.
\end{itemize}
Unfortunately, UGEs in training suffer from huge variance in learning with CLs.
Here, the \emph{root cause} of overfitting is that only one fixed CL is available for each instance, and the \emph{direct cause} is the huge variance of UGEs and the distance from empirical to true gradients.
The root cause also exists in other WSL problem settings, e.g., partial or noisy labels.
Notice that the quality of gradient estimation matters more than risk estimation in risk minimization, since stochastic optimizers mainly rely on empirical gradients.

Next, we propose a novel framework named \emph{surrogate complementary loss}~(SCL) to improve gradient estimation.
Recall that the \emph{classification error} is defined as the expected zero-one loss over the test distribution.
Existing URE-based methods first replace the zero-one loss with a surrogate loss to obtain the risk, and then rewrite the risk into an expectation over the training distribution.
We call it \emph{complementary surrogate loss} since replacing is before rewriting.
On the other hand, our framework first rewrites the error into an expectation over the training distribution and then replaces the zero-one loss with a surrogate loss, namely, rewriting before replacing.
Rewriting the error is nicer since the zero-one loss has many nice properties while the surrogate loss is just arbitrary.
In our experiments, SCL-based methods outperform URE-based methods, where SCL successfully reduces the variance of empirical gradients and makes them more aligned with true gradients in the direction. 

The rest of the paper is organized as follows.
We introduce WSL problem settings and the overfitting issue in Section \ref{s:ure}.
In Section \ref{s:prop}, we propose the SCL framework.
In Section \ref{s:grad}, we analyze empirical gradients to justify our claims.


\section{The Use of Unbiased Risk Estimators}
\label{s:ure}
In this section we introduce the usage of unbiased risk estimators in several weakly supervised learning settings.
Then we zoom into the problem of learning with complementary labels, and show the relationship between negative risk problem and overfitting.

\subsection{Related WSL Settings}
The following problems are typical examples where UREs fail under weak supervision.
The negative empirical risk can happen when the loss functions are not specifically restricted, causing overfitting.
Biased loss functions or non-negative correction methods are introduced to mitigate such issues in related literature.

\paragraph{Noisy Label Learning:}
Noisy label learning studies about learning when training labels flip according to some underlying distribution.
A common assumption is the class conditional noise setting where the noisy label depends on its ordinary label.
\citet{natarajan2013learning} first provided a URE for arbitrary loss in the binary case, and provided performance guarantee.
To ensure the convexity of the rewritten loss function, they require the original surrogate loss to satisfy a symmetric property.
\citet{patrini2017making} extends to multiclass classification and proposed two loss correction methods: backward correction and forward correction.
Backward correction involves a matrix inversion and gives an unbiased estimator of the original loss.
Forward correction corrects the prediction with a matrix multiplication and can be added as an additional layer to neural networks.
The authors showed that forward correction performs better than backward correction, and hinted the reason to be optimization related.

\paragraph{Positive-Unlabeled (PU) Learning:}
In binary classification, the labeled data consists of two sets, the positive (P) class and the negative (N) class.
PU learning studies when labeled data only consists of positive examples, while we have unlabeled (U) data consisting of both positive and negative examples.
\citet{elkan2008learning} proposed to learn from assigning weights to unlabeled examples.
\citet{du2014analysis} proposed a URE of non-convex losses, and \citet{du2015convex} extends it further to a more general framework with convex formulation.
\citet{kiryo2017positive} observed the overfitting issue of unbiased PU learning and proposed a non-negative risk estimator to fix the problem.

\paragraph{Unlabeled-Unlabeled (UU) learning:}
In binary classification, UU learning considers the setting when all labels are unknown.
\citet{lu2018minimal} discovers that if the two sets of data have different class priors, a URE can be derived to learn from such data.
However, the unbiased UU learning also encounters severe overfitting due to negative empirical risk.
\citet{lu2019mitigating} proposed a non-negative corrected risk estimator to fix the problem.

\subsection{Learning with Complementary Labels}

In the following part, we first introduce related work of learning with complementary labels, then formally define the URE formulation and the negative risk effect.

In \citet{ishida2017learning}, the first work to introduce the setting of complementary labels,
a URE can be obtained
when a loss function satisfies the symmetric property, under uniform assumptions. 
\citet{yu2018learning} provides a loss correction method for softmax cross
entropy loss, and shows that non-uniform complementary labels can also be
learned if the complementary transition matrix is known.
Continuing in the uniform complementary assumption of \citet{ishida2017learning}, \citet{ishida2019complementary} generalizes the URE for arbitrary loss functions
and models, and proposes a non-negative correction and a gradient ascent method to
account for overfitting.
Several studies have also extended to learning with multiple complementary 
labels~\cite{feng2019learning}, and its combination with unlabeled 
data~\cite{cao2020multi}.
The flexibility of CLs makes it easy to use in settings
such as online learning~\cite{kaneko2019online},  generative-discriminative
learning~\cite{xu2019generative}, and noisy label learning~\cite{kim2019nlnl}.

\paragraph{Ordinary Learning:}
We start by reviewing the setting and introduce notations in ordinary learning.
Consider the problem of $K$ class classification $(K>2)$, where $[K]=\{1, 2, ...,
K\}$ is the label set.
Let $D$ be a joint distribution over the feature set $X$ and label set $Y$,
where we sample input feature $x\in\mathbb{R}^d$ and label $y\in[K]$.
Given training samples $\{(x_i, y_i)\}_{i=1}^{n}$, the goal of the learning
algorithm is to learn a classifier $f(x):\mathbb{R}^d\rightarrow[K]$ which
predicts the correct label from a given input $x$.
The classifier $f$ is implemented with a decision function
$\bm{g}:\mathbb{R}^d\rightarrow\mathbb{R}^K$ by taking the argmax function
$f(x)=\argmax_i \bm{g}(x)_i$.
For a label $y$ and a decision function output $\bm{g}(x)$, the \emph{loss function} is defined as a nonnegative function $\ell:[K]\times\mathbb{R}^K\rightarrow\mathbb{R}^+$.
Finally, we define the \emph{risk} as the expected loss of $\bm{g}$ over distribution $D$:
\begin{equation}
    R(\bm{g};\ell)=\mathbb{E}_{(X,Y)\sim D}[\ell(Y, \bm{g}(X))].
    \label{eq:risk}
\end{equation}

\paragraph{Complementary Learning:}
In complementary learning, the data distribution is switched to
$\overline{D}=X\times\overline{Y}$ where the training samples given to the learner become $\{(x_i,
\overline{y}_i)\}_{i=1}^{n}$.
For instance $x_i$, the complementary label~(CL) $\overline{y}_i$ is a class in $[K]$ that $x_i$ does not belong to, satisfying $\overline{y}_i \neq y_i$.
In this case, the loss function $\ell$ cannot be used directly since the
ordinary target $y_i$ is not given.
In the following part, we review the derivation of URE using backward loss rewriting process~\cite{patrini2017making, ishida2019complementary}.

\paragraph{Unbiased Risk Estimator:}
In this part, we follow the assumption of class conditional complementary transition as in related work, assuming the transition matrix $T$ invertible, where $T_{ij}=\mathbb{P}(\overline{Y}=j\mid Y=i)$ and $T_{ii}=0$ for all $i$.
We borrow the following notation from \citet{ishida2019complementary}.
The loss vector is $\ell(\bm{g}(x))=[\ell(1, \bm{g}(x)), \ell(2, \bm{g}(x))...\ell(K, \bm{g}(x))]$, and let $e_{i}\in \{0,1\}^K$ denote the one-hot vector in which the $i$-th entry is one.
\begin{prop}
\label{p1}
The ordinary risk can be transformed as
\begin{equation}
    R(\bm{g};\ell)=\mathbb{E}_{(X, \overline{Y})\sim \overline{D}} [e_{\overline{y}}^\top (T^{-1}) \ell(\bm{g}(x))].
    \label{eq:matinv}
\end{equation}
That is, we obtain an unbiased risk estimator (URE):
\begin{equation}
    \overline{R}(\bm{g};\overline{\ell}) = \mathbb{E}_{(x,\overline{y})\sim\overline{D}}[\overline{\ell}(\overline{y}, \bm{g}(x))]
    =R(\bm{g};\ell)
    \label{eq:ure}
\end{equation}
where $\overline{\ell}$ is the following rewritten loss:
\begin{equation}
    \overline{\ell}(\overline{y}, \bm{g}(x)) = 
    e_{\overline{y}}^\top (T^{-1}) \ell(\bm{g}(x)).
\end{equation}
\end{prop}
This proposition implies the expectation of $\overline{\ell}(\overline{y}, \bm{g}(x))$ under distribution $\overline{D}$ is equivalent to the ordinary risk $R(\bm{g}; \ell)$.

\paragraph{Uniform Assumption:}
In the rest of this paper, we assume CLs are sampled uniformly from $[K]\setminus\{y\}$, for a better comparison with \citet{ishida2019complementary}.
By plugging in the uniform assumption  $T=\frac{1}{K-1}(\bm{1}_k-\bm{I}_k)$, 
we have the following formulation of $\overline{\ell}$,
\begin{equation}
    \overline{\ell}(\overline{y}, \bm{g}(x)) = -(K-1)\ell(\overline{y},\bm{g}(x)) + \sum_{j=1}^{K}\ell(j,\bm{g}(x)).
    \label{eq:ureloss}
\end{equation}
This URE approach minimizes $\overline{\ell}$ over the training distribution, and theoretical results from \citet{ishida2017learning} proved the consistency of the risk estimator under specific losses.

\subsection{Negative Risk and Overfitting}
However, URE tends to have poor empirical performance.
\citet{ishida2019complementary} reported that minimizing URE causes the empirical risk to go negative, which is a sign of overfitting.
It is clear that the negative loss term $-(K-1)\ell(\overline{y},\bm{g}(x))$ in $\overline{\ell}$ (Equation \ref{eq:ureloss}) is the source of negativity.
Such negative term occurs in common class conditional complementary transition as long as all diagonal elements of $T$ are zero.

Recall the URE in Equation \ref{eq:ure}, 
in expectation has minimum value 0
when the classifier has no error.
However, when minimizing URE empirically, the non-negative lower bound does not remain.
We claim that the main difference between the expectation and its empirical realization is the label distribution: only single $\overline{y}$ is given for each instance in practice, while the expectation is calculated over all possible $\overline{y}$.
The URE only stays non-negative when taken expectation, which is not realistic.

\paragraph{Negative Risk Experiment:}
To show the difference between theory (expectation) and practice,
we use an experiment to demonstrate how the empirical distribution of CLs leads to negative empirical risk during training.
Three different label distributions are given:
\begin{enumerate}
     \item Ordinary Learning (ORD): The supervised learning baseline, which the ordinary label $y$ is given. This is also the case where the complementary label is marginalized out by taking expectation.
     \item Fixed Complementary Learning (FIXED): The realistic complementary learning scenario, for each instance $x$ only a fixed CL $\overline{y}$ is given.
     \item Random Complementary Learning (RAND): The $\overline{y}$ of each instance is randomly sampled from $[K]\setminus\{y\}$ in each epoch. This setting acts as a stochastic version of ORD on $\overline{y}$.
\end{enumerate}
In this experiment, we used the cross-entropy loss as $\ell$ for ordinary learning
(ORD) and $\overline{\ell}$ for complementary learning (FIXED, RAND).
For MNIST, we use linear model and  single hidden
layer MLP $(d-500-10)$ as learning models; for CIFAR-10, we used ResNet-34~\cite{he2016deep} and DenseNet~\cite{huang2017densely}.
The models are trained with Adam~\cite{kingma2014adam} optimizer at a fixed learning rate of $10^{-5}$ for 300 epochs.

\begin{figure*}
    \centering
    \resizebox{\textwidth}{!}{
    \begin{tabular}{cccc}
    \includegraphics[width=0.33\textwidth]{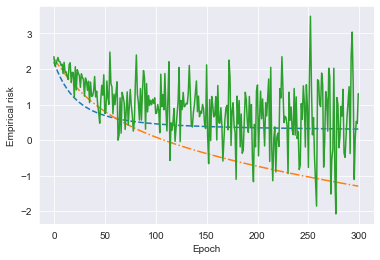} &
    \includegraphics[width=0.33\textwidth]{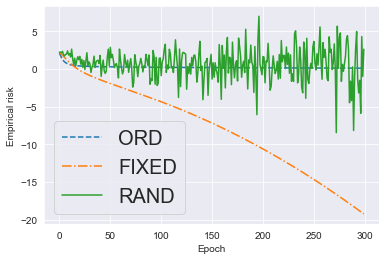} &
    \includegraphics[width=0.33\textwidth]{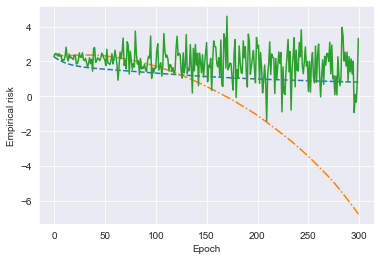} \\
    (a) MNIST, Linear & (b) MNIST, MLP & (c) CIFAR-10, DenseNet
 \end{tabular}
 }
\caption{Empirical risk minimization comparison}
\label{fig:risk_compare}
\end{figure*}

Results are shown in Figure~\ref{fig:risk_compare}.
FIXED suffers from severe negative risk in comparison to ORD and RAND, which is a clear sign of overfitting to the given CL.
The problem worsen as flexible models are used, matching results
from \citet{ishida2019complementary}.
However, note that RAND yields a significantly different result from
FIXED even though they are trained on the same objective.
Though the risk of RAND fluctuates considerably due to the changes in each epoch,
it does not stay negative, as we can view RAND as an randomized approximation of ORD.
The results also show that the estimated risk diverges far from the ordinary risk as the training goes on, and the gap increases with the training epochs.
In this case, consistency guarantees become ineffective since the risk estimation error keeps increasing as training goes on.
That is, the behavior of URE and the ordinary risk is extremely different in the empirical setting, even if statistical properties such as unbiasedness and consistency can be proven.

\paragraph{Risk Correction Methods:} 
\citet{ishida2019complementary} proposed two correction methods to mitigate the problem.
First, the non-negative loss correction (NN),
which enforces non-negativity to the decomposed risk of each class.
Second, namely the gradient ascent correction (GA) which enforces a reverse gradient update to the model parameters when the decomposed risk goes negative or under a certain threshold.
GA can be viewed as a more aggressive correction than NN.
The correction methods show improvements in various experiments, and similar techniques have also been applied in other WSL problems \cite{kiryo2017positive, lu2019mitigating}.
However, such correction methods are still based on URE and lack theoretical motivation, the fundamental difference between risk and URE are not solved.
We will include experiment results of these methods in the following sections.

\section{Proposed Framework}
\label{s:prop}
In this section, we propose a complementary learning framework that avoids the negative risk problem of URE.
To clearly distinguish between complementary learning and ordinary learning, we rethink the relationship between input features and labels:
An ordinary label provides a positive feedback to the given class, while a CL provides a negative feedback to the given class.
The maximum likelihood approach is commonly used in ordinary learning when we have probability estimation from the model, by maximizing the conditional likelihood given the training data.
The commonly used softmax cross-entropy loss function in deep learning is a typical example by combining softmax activation function and the maximum likelihood approach.
In complementary learning, given only CLs as training data, we propose to apply the minimum complementary likelihood approach, through a proxy loss.
In the following of this section, we propose a new framework that consists complementary 0-1 loss and its corresponding surrogate complementary loss (SCL).

\subsection{Complementary 0-1 Loss}
From the classification error perspective: 
In ordinary learning, zero error is obtained when the classifier predicts the correct class as the label, and has error otherwise.
In complementary learning, given only limited information, we can only be sure that prediction error occurs when the CL is predicted by the classifier.
With the rules above, we formally define the ordinary classification error and a novel complementary classification error: 
\begin{definition}
(Multiclass) classification error, or 0-1 loss:
\begin{equation}
    \ell_{01}(y, f(x))=\llbracket y\neq f(x)\rrbracket.
\end{equation}
\end{definition}

\begin{definition}
Complementary classification error, or complementary 0-1 loss:
\begin{equation}
    \overline{\ell}_{01}(\overline{y}, f(x))=\llbracket \overline{y}= f(x)\rrbracket.
\end{equation}
\end{definition}

$\overline{\ell}_{01}$ is 1 when the predicted class matches the CL, which indicates classification error.
By minimizing $\overline{\ell}_{01}$, we can minimize the conditional probability output of CLs.
\begin{prop}
\label{p2}
The complementary 0-1 loss is a constant multiple of the URE of the classification error.
\begin{equation}
R(\bm{g};\ell_{01}) 
= (K-1)\overline{R}(\bm{g};\overline{\ell}_{01})
\end{equation}
In other words, the URE of the classification error has the same minimizer with the complementary 0-1 loss:
\begin{equation}
    \mathbb{E}_{(x,\overline{y})\sim\overline{D}}
    [\overline{\ell}_{01}(\overline{y}, \bm{g}(x))]
\end{equation}
\end{prop}
Thus, existing guarantees show that we can learn with CLs via empirical risk minimization from
$\overline{R}(\bm{g};\overline{\ell}_{01})$.

\subsection{Surrogate Complementary Loss}
To minimize the non-convex $\overline{\ell}_{01}$, a common approach in statistical learning is to select a convex surrogate loss to approximate the target loss.
In order to minimize the output of the label prediction, which is the opposite of
most common surrogate functions, we require a new type of surrogate complementary loss (SCL) for
this problem setting.
Different from ordinary surrogate losses which are non-increasing functions of the label class output, SCLs are non-decreasing functions of the CL class output.

\paragraph{Baseline Methods:} 
To better distinguish from URE-based methods, we use $\phi$ to denote the SCL
loss functions.
Here we denote the probability output $\bm{p}\in\Delta^{K-1}$ if $\bm{g}$
passes through a softmax layer, where $\Delta^{K-1}$ is the $K$-dimensional simplex.
Existing work on complementary learning has resulted in similar patterns
that minimize label class prediction output.
We include these methods as baselines in our experiments.
\begin{enumerate}
     \item Forward correction~(SCL-FWD) in  \citet{yu2018learning}: a forward loss correction method given transition matrix $T$:
    \begin{equation}
        \phi_{\text{FWD}}(\overline{y}, \bm{g}(x)) = \ell(\overline{y}, T^\top \bm{p}).
    \end{equation}
    
     \item Negative learning loss~(SCL-NL) in \citet{kim2019nlnl}: a modified log loss for negative learning with CLs:
    \begin{equation}
        \phi_{\text{NL}}(\overline{y}, \bm{g}(x)) = 
        -\log(1 - \bm{p}_{\overline{y}}).
    \end{equation}
    
    \item Exponential loss~(SCL-EXP):
    \begin{equation}
        \phi_{\text{EXP}}(\overline{y}, \bm{g}(x)) = 
        \exp(\bm{p}_{\overline{y}}).
    \end{equation}
\end{enumerate}

As we unify the above-mentioned losses into the surrogate complementary loss
$\phi$ framework.
These loss functions actually all accomplish the same purpose: minimizing the
complementary 0-1 loss by using its loss as surrogate:
\begin{equation}
    \min\overline{\ell}_{01}(\overline{y}, f(x)) \rightarrow
    \min\phi(\overline{y}, \bm{g}(x)).
\end{equation}

\begin{figure*}[ht]
    \centering
    \begin{tikzpicture}[auto, node distance=2cm,>=latex']
        \node [] at (-1,1) (ure) {URE};
        \node [] at (-1,-1) (csl) {SCL};
        
        \node [draw] at (0,0) (risk_01)
        {\Large$R_{01}$};
        \node [draw] at (4,1) (risk_l) {\Large$R_{\ell}$};
        \node [draw, right of=risk_l, node distance=4cm] (cr_cl) {\Large$\overline{R}_{\overline{\ell}}$};
        \node [draw, right of=cr_cl, node distance=3.5cm] (h_cr_cl) {\Large$\hat{\overline{R}}_{\overline{\ell}}$};            
        \node [draw] at (4,-1) (cr_c01) {\Large$\overline{R}_{\overline{01}}$};
        \node [draw, right of=cr_c01, node distance=4cm] (cr_phi) {\Large$\overline{R}_{\phi}$};
        \node [draw, right of=cr_phi, node distance=3.5cm] (h_cr_phi) {\Large$\hat{\overline{R}}_{\phi}$};
        
        \draw [->] (risk_01) -- node [sloped, pos=0.8] {Approximation} (risk_l);
        \draw [->] (risk_l) -- node {Estimation} (cr_cl);
        \draw [->] (cr_cl) -- node {Empirical} (h_cr_cl);
        
        \draw [->] (risk_01) -- node[below, sloped] {Estimation} (cr_c01);
        \draw [->] (cr_c01) -- node[below] {Approximation} (cr_phi);
        \draw [->] (cr_phi) -- node[below] {Empirical} (h_cr_phi);
    \end{tikzpicture}
    \caption{Comparison of URE learning process with the SCL framework.}
    \label{fig:framework}
\end{figure*}
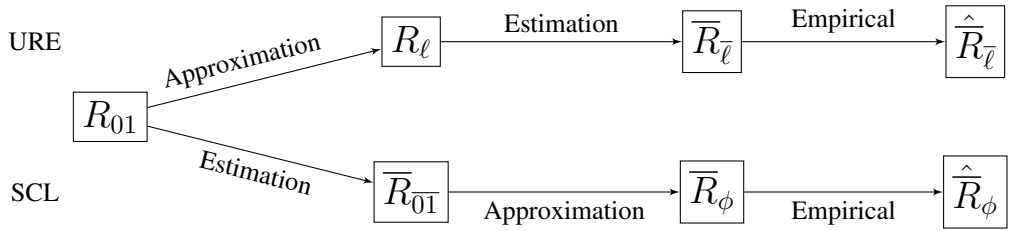

Here we compare the proposed SCL learning process with the URE learning
process, as shown in Figure~\ref{fig:framework}.
We use \emph{approximation} step to denote the process of replacing 0-1 loss with its surrogate loss, and the \emph{estimation} step represents rewriting the risk from ordinary distribution to complementary distribution.
Given the same goal of minimizing the true classification risk $R_{01}$, the
two frameworks follow a different order in the learning steps.
The URE framework follows the traditional statistical learning framework by
approximating $R_{01}$ with $R_{\ell}$, and then performs the estimation step by
rewriting the risk into $\overline{R}_{\overline{\ell}}$ for the complementary
distribution.
The SCL framework, on the other hand, performs the approximation step after the estimation step by first rewriting the classification risk $R_{01}$ to complementary
classification risk $\overline{R}_{\overline{01}}$, 
then perform the approximation step by using the SCL loss
$\phi$, resulting in the objective $\overline{R}_{\phi}$.

The ordinary surrogate loss $\ell$ in URE is used for ordinary labels, which serves as
an upper bound proxy in order to minimize the 0-1 classification error.
However, when the training data distribution is changed into CLs, the loss is rewritten and the non-negativity of $\ell$ no longer remain, causing the negative risk term.
That is, a ripple effect of error happens when the approximation error of the surrogate loss is amplified by the estimation step.
In the proposed SCL framework, we sidestep this question by placing the surrogate 
process after the risk rewriting.
In this way, the surrogate loss $\phi$ is directly applied on its target $\overline{\ell}_{01}$, and the negative loss problem is avoided.
Furthermore, it is not only the statistical properties that matters to surrogate loss, optimization properties such as smoothness and curvature are also important to
consider.
As the estimation process of URE damages the original properties of $\ell$, the optimization properties of $\phi$ are preserved.

\begin{table*}[ht]
\caption{Classification accuracies}
\label{acc-table}
\vskip 0.15in
\begin{center}
\begin{small}
\begin{sc}
\begin{tabular}{lc|cc|ccccr}
\toprule
data set + Model & URE & NN & GA & SCL-FWD & SCL-NL & SCL-Exp \\
\midrule
MNIST + Linear & 0.8503 & 0.8182 & 0.8193 & 0.9 & 0.9 & \textbf{0.9019} \\
MNIST + MLP & 0.8012 & 0.8665 & 0.9088 & 0.8965 & \textbf{0.9469} & 0.9251 \\
Kuzushi-MNIST + Linear & 0.5613 & 0.5331 & 0.4992 & 0.6056 & 0.6056 & \textbf{0.6132} \\
Kuzushi-MNIST + MLP & 0.5433 & 0.5683 & 0.6567 & 0.6445 & \textbf{0.7644} & 0.7184 \\
Fashion-MNIST + Linear & 0.7675 & 0.7755 & 0.7672 & 0.8274 & 0.8274 & \textbf{0.8282} \\
Fashion-MNIST + MLP & 0.7401 & 0.7829 & 0.8019 & 0.8372 & \textbf{0.8456} & 0.835 \\
CIFAR-10 + Resnet & 0.1091 & 0.3078 & 0.3738 & \textbf{0.5058} & 0.4713 & 0.492 \\
CIFAR-10 + Densenet & 0.2909 & 0.3379 & 0.4108 & \textbf{0.5457} & 0.5394 & 0.5435 \\
\bottomrule
\end{tabular}
\end{sc}
\end{small}
\end{center}
\vskip -0.1in
\end{table*}

\subsection{Classification Accuracy}
In this section, we use an experiment to compare the performance of each
method.
Specifically, the methods can be classified into two categories: URE-based methods, and
SCL-based methods. In URE-based methods, we have URE, URE with negative risk correction (NN), and URE with gradient ascent (GA). In SCL-based methods, we have SCL-FWD, SCL-NL, and SCL-EXP.
We used the Adam optimizer with learning rate selected from
$\{10^{-1}, 10^{-2}, 10^{-3}, 10^{-4}, 10^{-5}\}$ and trained the models for
300 epochs.

The testing accuracy is shown in Table~\ref{acc-table}.
The URE performs poorly compared to other methods, especially in more flexible models.
Even though NN and GA improve on URE in most tasks, the SCL
methods still outperform them by a significant gap.
These results justify our claims.
Although URE is an estimation of the risk~$R_{\ell}$ with statistical guarantees, in practice, it does not perform well as a classifier.     
On the other hand, although the proposed SCL framework is biased to the risk
$R_{\ell}$, introducing such bias towards minimizing the CL output yields superior results compared to URE, avoiding the negative risk issue.
In the next section, we discuss why the difference between the two
frameworks result in such a performance gap by analyzing the loss gradient
during training.

\newcommand{\vf}{\bm{f}}
\newcommand{\vc}{\bm{c}}
\newcommand{\vh}{\bm{h}}

\newcommand{\va}{\bm{a}}
\newcommand{\vb}{\bm{b}}
\newcommand{\gl}{\nabla_{\theta}\ell(y, \bm{g}(x))}
\newcommand{\glc}{\nabla_{\theta}\overline{\ell}(\overline{y}, \bm{g}(x))}
\newcommand{\gphi}{\nabla_{\theta}\phi(\overline{y}, \bm{g}(x))}
\newcommand{\cy}{\overline{y}}

\section{Gradient Analysis}
\label{s:grad}
In this section, we discuss how the proposed SCL framework outperforms URE through two gradient analysis experiments.
As mentioned in Section~\ref{s:ure}, the URE diverges widely from the risk itself when only a single CL is used to estimate the risk.
Here we further discuss how the SCL framework gives such improvement by rearranging the learning process.
The discussion will focus on the loss gradient: in the experiments, they are the
stochastic gradient (SGD) in mini-batch optimization specifically.
The analysis can be viewed as two parts: gradient directional estimation,
and the bias-variance tradeoff of the gradient estimation error.

\begin{figure*}
  \centering
  \resizebox{\textwidth}{!}{
  \begin{tabular}{cccc}
    \includegraphics[width=0.33\textwidth]{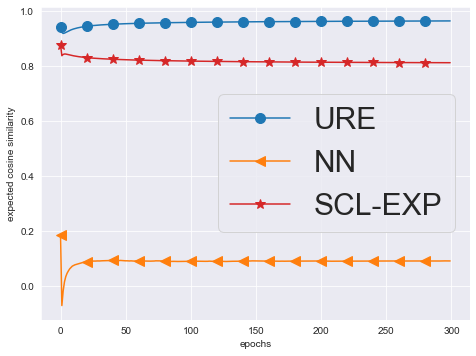} &
    \includegraphics[width=0.33\textwidth]{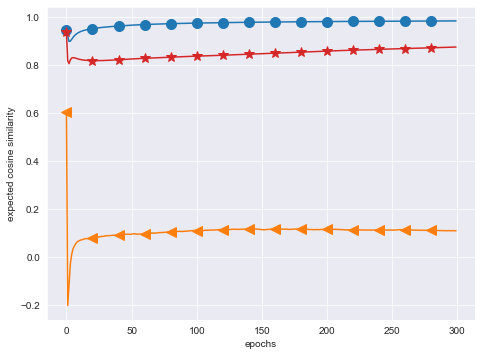} &
    \includegraphics[width=0.33\textwidth]{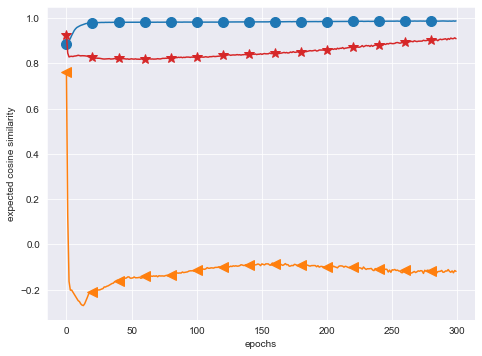}\\
    
    (a-1) Expected: MNIST, Linear & (b-1) Expected: MNIST, MLP & (c-1) Expected: CIFAR-10, DenseNet
    \\
    
    \includegraphics[width=0.33\textwidth]{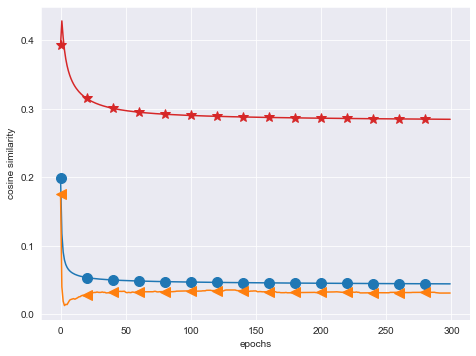} &
    \includegraphics[width=0.33\textwidth]{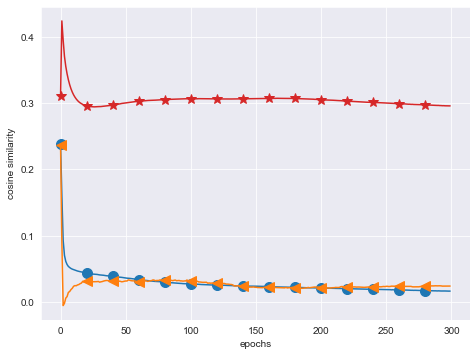} &
    \includegraphics[width=0.33\textwidth]{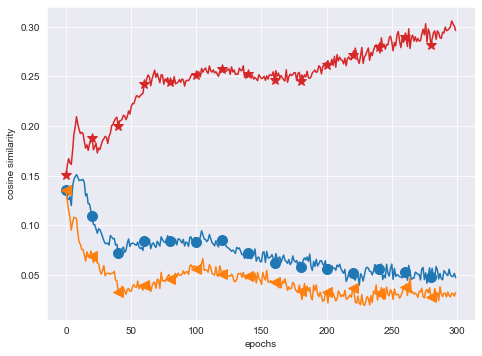}\\
    
    (a-2) Fixed: MNIST, Linear & (b-2) Fixed: MNIST, MLP & (c-2) Fixed: CIFAR-10, DenseNet
 \end{tabular}
 }
\caption{Cosine similarity comparison.}
\label{fig:cosinesim}
\end{figure*}

    
    
    

\subsection{Directional Similarity}
\label{ss:dirsim}
Since the URE is an estimator of the risk function, we expect its optimization
to be similar to the risk function.
Here we prove that the gradient of URE is also an unbiased gradient estimator~(UGE) of the ordinary gradient.
\begin{prop}
\label{p3}
The gradient of an unbiased risk estimator is unbiased to the ordinary risk gradient.
That is, for an instance $(x, y)$ we have,
\begin{equation}
    \mathbb{E}_{\overline{y}\mid y}
    \big[
    \nabla_{\theta}\overline{\ell}(\overline{y}, \bm{g}(x))
    \big] = \nabla_{\theta}\ell(y, \bm{g}(x))
\end{equation}
\end{prop}
Thus, the gradient of the complementary loss $\overline{\ell}$ is unbiased
with respect to the gradient of the ordinary loss, in our case the gradient of cross
entropy loss.
However, does that lead to similar performance with ordinary learning?
Our experimental results show that URE methods learn poorly through unbiased
gradient estimation.

In this section, we use an experiment to compare the gradient direction of
ordinary learning and its complementary learning counterparts.
We compare the complementary loss gradient directions with the ordinary gradient direction of the cross
entropy loss $\gl=-\nabla_{\theta}\log(\bm{p}_y)$.

The quality of the complementary gradient depends on its similarity with the ordinary gradient direction, where the similarity of two gradient directions is measured by the cosine similarity $\mathbb{S}$ of two gradient vectors $\bm{a}$ and $\bm{b}$, where $\mathbb{S}(\bm{a}, \bm{b}) = (\bm{a}\cdot\bm{b})/|\bm{a}||\bm{b}|$.
For the gradient directions, a reasonable assumption is $\mathbb{S}$ should be as large
as possible, indicating a direction more similar to the ordinary gradient.
In this experiment, we compare two gradient settings:
\begin{enumerate}
     \item Expected: The averaged gradient computed over all possible CLs of an instance $x$.
     \item Fixed: The gradient computed on a single CL of an instance $x$.
\end{enumerate}

We compared three complementary learning methods on their
approximation of the direction of the ordinary gradient: URE, NN, SCL-EXP.
To ensure fair comparison, the model is
updated only with ordinary labels in each epoch to avoid gradient error
accumulation, the complementary gradients were computed only for comparison and were not
updated to the model.
The SGD optimizer was used with a learning rate fixed at $10^{-2}$, trained for 300
epochs.

As the results show in Figure~\ref{fig:cosinesim}, URE achieves an ideal
gradient direction only in the case of expected CLs.
In the fixed case, UGE results in very different gradient directions with respect to the
ordinary gradient direction.
This shows that in the case when each $x$ is fixed to a $\overline{y}$, UGE
does not estimate a reliable direction.
The UGEs of each $\overline{y}$ have diverged directions in order to maintain the unbiasedness.
Unsurprisingly, the SCL methods provide better approximations of the ordinary gradient,
since it does not diverge by focusing on the CL direction $\overline{\ell}_{01}$.


\begin{figure*}[ht]
  \centering
  \resizebox{\textwidth}{!}{
  \begin{tabular}{ccc}
    \includegraphics[width=0.33\textwidth]{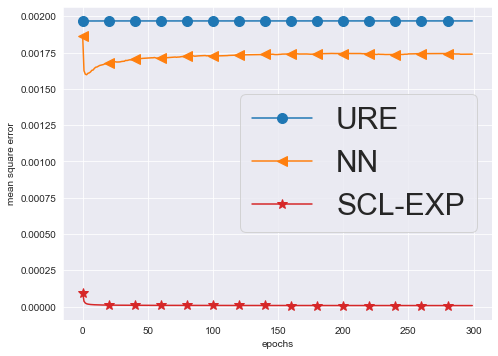} &
    \includegraphics[width=0.33\textwidth]{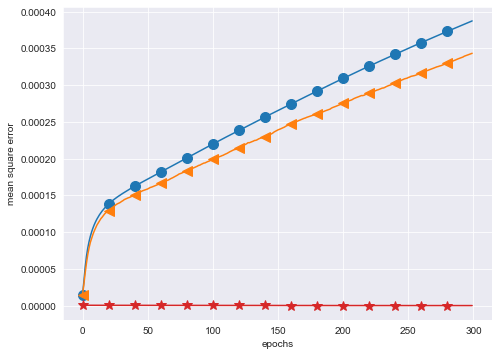} &
    \includegraphics[width=0.33\textwidth]{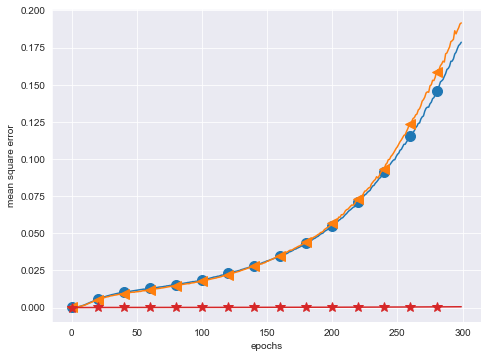}
    \\
    (a-1) MSE: MNIST, Linear & (b-1) MSE: MNIST, MLP & (c-1) MSE: CIFAR-10, DenseNet \\

    \includegraphics[width=0.33\textwidth]{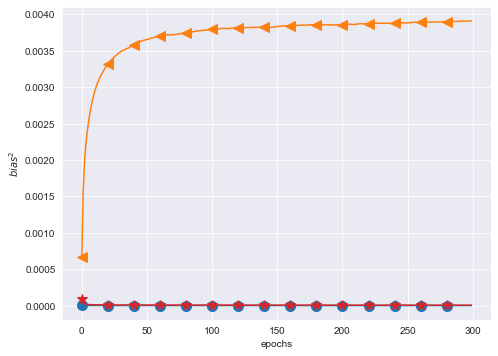} &
    \includegraphics[width=0.33\textwidth]{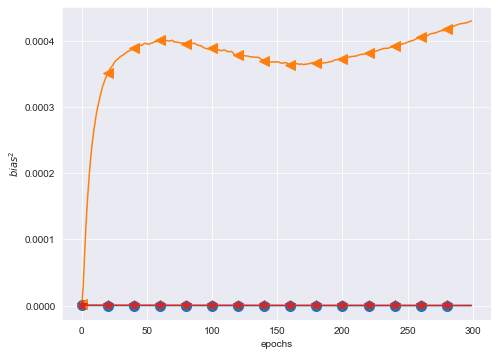} &
    \includegraphics[width=0.33\textwidth]{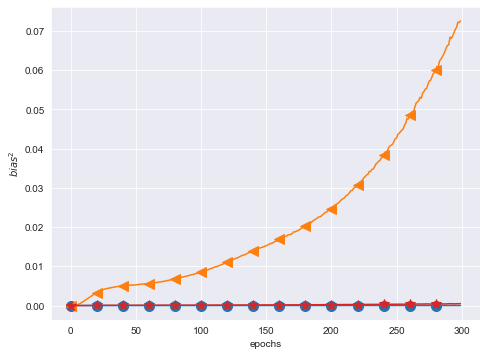}
    \\
    (a-2) $\text{Bias}^2$: MNIST, Linear & (b-2) $\text{Bias}^2$: MNIST, MLP & (c-2) $\text{Bias}^2$: CIFAR-10, DenseNet \\
    
    \includegraphics[width=0.33\textwidth]{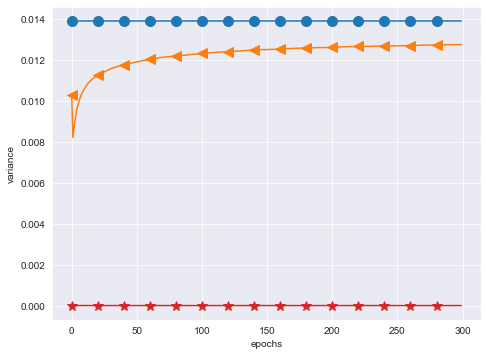} &
    \includegraphics[width=0.33\textwidth]{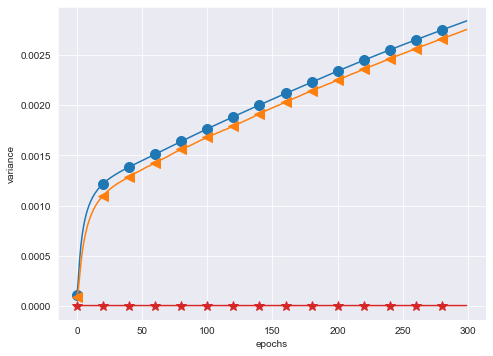} &
    \includegraphics[width=0.33\textwidth]{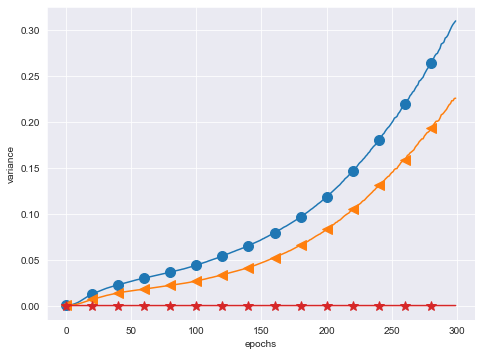}
    \\
    (a-3) Variance: MNIST, Linear & (b-3) Variance: MNIST, MLP & (c-3) Variance: CIFAR-10, DenseNet
    
 \end{tabular}
 }
\caption{Error decomposition of gradient estimators.}
\label{fig:msecompare}
\end{figure*}

\subsection{Bias-Variance Tradeoff}
In this part, we further analyze the estimation error of the complementary gradient verses the ordinary gradient, using the bias-variance decomposition technique.
Bias-variance decomposition is a common approach in statistical learning used to evaluate the complexity of a learning algorithm; instead of analyzing the error of a prediction problem, we extend this framework to evaluate the estimation error of the gradient, setting the ordinary gradient as the target.
We will show that URE has much larger $L_2$ loss than SCL caused by its large variance, despite having no bias.

We denote $\vf$ as the gradient step determined by ordinary labeled data $(x, y)$ and ordinary loss $\ell$.
$\vc$ denotes the complementary gradient step by complementary labeled data $(x, \overline{y})$ and complementary loss $\overline{\ell}$ (or $\phi$).
$\vh$ denotes the expected gradient step of $[K]\setminus\{y\}$, which is the average of $\vc$ on every possible CL.
We formalize as:
\begin{align}
    \vf &= \nabla\ell(y, \bm{g}(x)) \\
    \vc &= \nabla\overline{\ell}(\overline{y}, \bm{g}(x)) \label{eq:vc} \\
    \vh &= \frac{1}{K-1} \sum_{y'\neq y} \nabla\overline{\ell}(y', \bm{g}(x))
    \label{eq:vh}
\end{align}
In this setting, we set $\vf$ as the ground truth, which is the target for the complementary estimator $\vc$.
We hope the mean squared error(MSE) of gradient estimation to be small.
\begin{equation}
    \text{MSE} = \mathbb{E}_{x, y, \overline{y}}\big[(\vf - \vc)^2 \big]
    \label{eq:mse}
\end{equation}

Here we can derive the bias-variance decomposition by introducing $\vh$ and eliminating remaining terms:
\begin{align}
\mathbb{E}\big[(\vf-\vc)^2 \big]
&= \mathbb{E}\big[(\vf-\vh+\vh-\vc)^2 \big] \\
&= \underbrace{\mathbb{E}\big[(\vf-\vh)^2 \big]}_{\text{Bias}^2} + \underbrace{\mathbb{E}\big[(\vh-\vc)^2\big]}_\text{Variance}
\label{eq:bv}
\end{align}
Since the UGE has no bias, it implies that all the estimation error of UGE comes from the variance term.

We run experiments to check how the complementary gradient $\vc$ approximate the ordinary gradient $\vf$, and compare with baseline methods.
The training works as follows.
In each epoch, we compute three gradients: the ordinary gradient $\vf$, the current method $\vc$, and $\vh$. 
We measure the mean square error (MSE), the squared bias term and the variance term according to Equation \ref{eq:mse} and Equation \ref{eq:bv}.
In each epoch, we only update the model with $\vf$ to maintain a fair comparison of the gradients.
The optimizer is SGD with a learning rate fixed at $10^{-2}$, trained for 300 epochs.

Results are showed in Figure \ref{fig:msecompare} (mean statistics are shown in Table \ref{mse_table1} and Table \ref{mse_table2}), GA is omitted for visualization reasons.
It is clear that although URE has no bias, it has very large MSE due to the large variance.
On the other hand, the SCL methods though have little bias, have much smaller variance compared to URE.
This justifies our claims in Section \ref{ss:dirsim}, the URE creates highly diverged gradients in order to maintain the overall unbiasedness, resulting high gradient variance.
On the other hand, SCL introduces \emph{inductive bias} towards minimizing the CL likelihood, trading zero bias with reduced variance.

\section{Conclusion}
In this paper, we show that unbiased risk estimator (URE) does not serve as a desirable optimization objective in weakly supervised learning problems such as learning with complementary labels.
From the empirical risk aspect, the URE encounters the negative risk issue which leads to severe overfitting under weakly supervision.
From the gradient aspect, the effort to maintain the unbiased gradient estimator (UGE) causes misleading direction and large variance to the loss gradient.
We propose a new SCL learning framework based on the minimum likelihood principle and surrogate complementary loss functions.
Though having a bias towards the CL, the SCL framework avoids the extremely noisy gradient problem encountered in URE.
Empirical results show that SCL outperforms URE in classification accuracy and other gradient quality metrics.

\begin{table}[ht]
\caption{Gradient error decomposition of MNIST on linear model\\
(Averaged over 300 epochs)}
\label{mse_table1}
\vskip 0.15in
\begin{center}
\begin{small}
\begin{sc}
\begin{tabular}{lcccr}
\toprule
Method & MSE & $\text{bias}^2$ & Variance \\
\midrule
URE & 1.9692E-03 & \textbf{8.0643E-07} & 1.3907E-02 \\
NN & 1.7268E-03 & 3.7248E-03 & 1.2272E-02 \\
GA & 1.0436E+00 & 2.5829E+00 & 8.3596E+00 \\
SCL-FWD & \textbf{7.7511E-06} & 7.5037E-06 & 6.9942E-07 \\
SCL-NL & \textbf{7.7511E-06} & 7.5038E-06 & 6.9931E-07 \\
SCL-EXP & 7.9152E-06 & 7.7895E-06 & \textbf{4.3945E-07} \\
\bottomrule
\end{tabular}
\end{sc}
\end{small}
\end{center}
\vskip -0.1in
\end{table}

\begin{table}[ht]
\caption{Gradient error decomposition of CIFAR-10 on DenseNet\\
(Averaged over 300 epochs)}
\label{mse_table2}
\vskip 0.15in
\begin{center}
\begin{small}
\begin{sc}
\begin{tabular}{lcccr}
\toprule
Method & MSE & $\text{bias}^2$ & Variance \\
\midrule
URE & 5.0196E-02 & \textbf{6.6855E-06} & 1.0101E-01 \\
NN & 5.2152E-02 & 2.1846E-02 & 7.0500E-02 \\
GA & 3.1350E+01 & 1.2985E+01 & 3.8302E+01 \\
SCL-FWD & \textbf{2.0237E-04} & 1.9225E-04 & 1.1051E-05 \\
SCL-NL & \textbf{2.0237E-04} & 1.9225E-04 & 1.1050E-05 \\
SCL-EXP & 2.0455E-04 & 1.9810E-04 & \textbf{7.0735E-06} \\
\bottomrule
\end{tabular}
\end{sc}
\end{small}
\end{center}
\vskip -0.1in
\end{table}


\section*{Acknowledgements}
GN and MS were supported by JST AIP Acceleration Research Grant Number JPMJCR20U3, Japan. YC and HL were partially supported by MOST 107-2628-E-002-008-MY3 and 108-2119-M-007-010.


\bibliography{paper}

\begin{thebibliography}{35}
\providecommand{\natexlab}[1]{#1}
\providecommand{\url}[1]{\texttt{#1}}
\expandafter\ifx\csname urlstyle\endcsname\relax
  \providecommand{\doi}[1]{doi: #1}\else
  \providecommand{\doi}{doi: \begingroup \urlstyle{rm}\Url}\fi

\bibitem[Bao et~al.(2018)Bao, Niu, and Sugiyama]{bao2018classification}
Bao, H., Niu, G., and Sugiyama, M.
\newblock Classification from pairwise similarity and unlabeled data.
\newblock In \emph{ICML}, 2018.

\bibitem[Cao \& Xu(2020)Cao and Xu]{cao2020multi}
Cao, Y. and Xu, Y.
\newblock Multi-complementary and unlabeled learning for arbitrary losses and
  models.
\newblock \emph{CoRR}, abs/2001.04243, 2020.
\newblock URL \url{https://arxiv.org/abs/2001.04243}.

\bibitem[Chapelle et~al.(2009)Chapelle, Scholkopf, and Zien]{chapelle2009semi}
Chapelle, O., Scholkopf, B., and Zien, A.
\newblock Semi-supervised learning (chapelle, o. et al., eds.; 2006)[book
  reviews].
\newblock \emph{IEEE Transactions on Neural Networks}, 20\penalty0
  (3):\penalty0 542--542, 2009.

\bibitem[du~Plessis et~al.(2015)du~Plessis, Niu, and Sugiyama]{du2015convex}
du~Plessis, M., Niu, G., and Sugiyama, M.
\newblock Convex formulation for learning from positive and unlabeled data.
\newblock In \emph{ICML}, pp.\  1386--1394, 2015.

\bibitem[du~Plessis et~al.(2014)du~Plessis, Niu, and Sugiyama]{du2014analysis}
du~Plessis, M.~C., Niu, G., and Sugiyama, M.
\newblock Analysis of learning from positive and unlabeled data.
\newblock In \emph{NeurIPS}, 2014.

\bibitem[Elkan \& Noto(2008)Elkan and Noto]{elkan2008learning}
Elkan, C. and Noto, K.
\newblock Learning classifiers from only positive and unlabeled data.
\newblock In \emph{KDD}, 2008.

\bibitem[Feng et~al.(2020)Feng, Kaneko, Han, Niu, An, and
  Sugiyama]{feng2019learning}
Feng, L., Kaneko, T., Han, B., Niu, G., An, B., and Sugiyama, M.
\newblock Learning with multiple complementary labels.
\newblock In \emph{ICML}, 2020.

\bibitem[Han et~al.(2018{\natexlab{a}})Han, Yao, Niu, Zhou, Tsang, Zhang, and
  Sugiyama]{han2018masking}
Han, B., Yao, J., Niu, G., Zhou, M., Tsang, I., Zhang, Y., and Sugiyama, M.
\newblock Masking: A new perspective of noisy supervision.
\newblock In \emph{NeurIPS}, 2018{\natexlab{a}}.

\bibitem[Han et~al.(2018{\natexlab{b}})Han, Yao, Yu, Niu, Xu, Hu, Tsang, and
  Sugiyama]{han2018coteaching}
Han, B., Yao, Q., Yu, X., Niu, G., Xu, M., Hu, W., Tsang, I., and Sugiyama, M.
\newblock Co-teaching: Robust training of deep neural networks with extremely
  noisy labels.
\newblock In \emph{NeurIPS}, 2018{\natexlab{b}}.

\bibitem[He et~al.(2016)He, Zhang, Ren, and Sun]{he2016deep}
He, K., Zhang, X., Ren, S., and Sun, J.
\newblock Deep residual learning for image recognition.
\newblock In \emph{CVPR}, 2016.

\bibitem[Hsieh et~al.(2019)Hsieh, Niu, and Sugiyama]{hsieh2019classification}
Hsieh, Y.-G., Niu, G., and Sugiyama, M.
\newblock Classification from positive, unlabeled and biased negative data.
\newblock In \emph{ICML}, 2019.

\bibitem[Huang et~al.(2017)Huang, Liu, Van Der~Maaten, and
  Weinberger]{huang2017densely}
Huang, G., Liu, Z., Van Der~Maaten, L., and Weinberger, K.~Q.
\newblock Densely connected convolutional networks.
\newblock In \emph{CVPR}, 2017.

\bibitem[Ishida et~al.(2017)Ishida, Niu, Hu, and Sugiyama]{ishida2017learning}
Ishida, T., Niu, G., Hu, W., and Sugiyama, M.
\newblock Learning from complementary labels.
\newblock In \emph{NeurIPS}, 2017.

\bibitem[Ishida et~al.(2018)Ishida, Niu, and Sugiyama]{ishida2018binary}
Ishida, T., Niu, G., and Sugiyama, M.
\newblock Binary classification from positive-confidence data.
\newblock In \emph{NeurIPS}, 2018.

\bibitem[Ishida et~al.(2019)Ishida, Niu, Menon, and
  Sugiyama]{ishida2019complementary}
Ishida, T., Niu, G., Menon, A., and Sugiyama, M.
\newblock Complementary-label learning for arbitrary losses and models.
\newblock In \emph{ICML}, 2019.

\bibitem[Jin \& Ghahramani(2002)Jin and Ghahramani]{jin2002partial}
Jin, R. and Ghahramani, Z.
\newblock Learning with multiple labels.
\newblock In \emph{NeurIPS}, 2002.

\bibitem[Kaneko et~al.(2019)Kaneko, Sato, and Sugiyama]{kaneko2019online}
Kaneko, T., Sato, I., and Sugiyama, M.
\newblock Online multiclass classification based on prediction margin for
  partial feedback.
\newblock \emph{arXiv preprint arXiv:1902.01056}, 2019.

\bibitem[Kim et~al.(2019)Kim, Yim, Yun, and Kim]{kim2019nlnl}
Kim, Y., Yim, J., Yun, J., and Kim, J.
\newblock Nlnl: Negative learning for noisy labels.
\newblock In \emph{ICCV}, 2019.

\bibitem[Kingma \& Ba(2015)Kingma and Ba]{kingma2014adam}
Kingma, D.~P. and Ba, J.
\newblock Adam: A method for stochastic optimization.
\newblock 2015.

\bibitem[Kiryo et~al.(2017)Kiryo, Niu, du~Plessis, and
  Sugiyama]{kiryo2017positive}
Kiryo, R., Niu, G., du~Plessis, M.~C., and Sugiyama, M.
\newblock Positive-unlabeled learning with non-negative risk estimator.
\newblock In \emph{NeurIPS}, 2017.

\bibitem[Lu et~al.(2019)Lu, Niu, Menon, and Sugiyama]{lu2018minimal}
Lu, N., Niu, G., Menon, A.~K., and Sugiyama, M.
\newblock On the minimal supervision for training any binary classifier from
  only unlabeled data.
\newblock In \emph{ICLR}, 2019.

\bibitem[Lu et~al.(2020)Lu, Zhang, Niu, and Sugiyama]{lu2019mitigating}
Lu, N., Zhang, T., Niu, G., and Sugiyama, M.
\newblock Mitigating overfitting in supervised classification from two
  unlabeled datasets: A consistent risk correction approach.
\newblock In \emph{AISTATS}, 2020.

\bibitem[Nagarajan \& Kolter(2019)Nagarajan and Kolter]{nagarajan2019uniform}
Nagarajan, V. and Kolter, J.~Z.
\newblock Uniform convergence may be unable to explain generalization in deep
  learning.
\newblock In \emph{NeurIPS}, 2019.

\bibitem[Natarajan et~al.(2013)Natarajan, Dhillon, Ravikumar, and
  Tewari]{natarajan2013learning}
Natarajan, N., Dhillon, I.~S., Ravikumar, P.~K., and Tewari, A.
\newblock Learning with noisy labels.
\newblock In \emph{NeurIPS}, 2013.

\bibitem[Niu et~al.(2016)Niu, du~Plessis, Sakai, Ma, and
  Sugiyama]{niu2016theoretical}
Niu, G., du~Plessis, M.~C., Sakai, T., Ma, Y., and Sugiyama, M.
\newblock Theoretical comparisons of positive-unlabeled learning against
  positive-negative learning.
\newblock In \emph{NeurIPS}, 2016.

\bibitem[Patrini et~al.(2017)Patrini, Rozza, Krishna~Menon, Nock, and
  Qu]{patrini2017making}
Patrini, G., Rozza, A., Krishna~Menon, A., Nock, R., and Qu, L.
\newblock Making deep neural networks robust to label noise: A loss correction
  approach.
\newblock In \emph{ICCV}, 2017.

\bibitem[Sakai et~al.(2017)Sakai, du~Plessis, Niu, and
  Sugiyama]{pmlr-v70-sakai17a}
Sakai, T., du~Plessis, M.~C., Niu, G., and Sugiyama, M.
\newblock Semi-supervised classification based on classification from positive
  and unlabeled data.
\newblock In \emph{ICML}, 2017.

\bibitem[Sakai et~al.(2018)Sakai, Niu, and Sugiyama]{sakai2018semi}
Sakai, T., Niu, G., and Sugiyama, M.
\newblock Semi-supervised auc optimization based on positive-unlabeled
  learning.
\newblock \emph{Machine Learning}, 107\penalty0 (4):\penalty0 767--794, 2018.

\bibitem[Vapnik(1992)]{vapnik1992principles}
Vapnik, V.
\newblock Principles of risk minimization for learning theory.
\newblock In \emph{NeurIPS}, 1992.

\bibitem[Xia et~al.(2019)Xia, Liu, Wang, Han, Gong, Niu, and
  Sugiyama]{xia2019anchor}
Xia, X., Liu, T., Wang, N., Han, B., Gong, C., Niu, G., and Sugiyama, M.
\newblock Are anchor points really indispensable in label-noise learning?
\newblock In \emph{NeurIPS}, 2019.

\bibitem[Xu et~al.(2020)Xu, Gong, Chen, Liu, Zhang, and
  Batmanghelich]{xu2019generative}
Xu, Y., Gong, M., Chen, J., Liu, T., Zhang, K., and Batmanghelich, K.
\newblock Generative-discriminative complementary learning.
\newblock 2020.

\bibitem[Yu et~al.(2018)Yu, Liu, Gong, and Tao]{yu2018learning}
Yu, X., Liu, T., Gong, M., and Tao, D.
\newblock Learning with biased complementary labels.
\newblock In \emph{ECCV}, 2018.

\bibitem[Yu et~al.(2019)Yu, Han, Yao, Niu, Tsang, and Sugiyama]{yu2019coplus}
Yu, X., Han, B., Yao, J., Niu, G., Tsang, I.~W., and Sugiyama, M.
\newblock How does disagreement help generalization against label corruption?
\newblock In \emph{ICML}, 2019.

\bibitem[Zhang et~al.(2017)Zhang, Bengio, Hardt, Recht, and
  Vinyals]{zhang2016understanding}
Zhang, C., Bengio, S., Hardt, M., Recht, B., and Vinyals, O.
\newblock Understanding deep learning requires rethinking generalization.
\newblock In \emph{ICLR}, 2017.

\bibitem[Zhou(2017)]{zhou2017brief}
Zhou, Z.-H.
\newblock A brief introduction to weakly supervised learning.
\newblock \emph{National Science Review}, 5\penalty0 (1):\penalty0 44--53,
  2017.

\end{thebibliography}
\bibliographystyle{icml2020}


\newpage
\onecolumn

\icmltitle{Supplementary Material for Unbiased Risk Estimators Can Mislead:\\
    A Case Study of Learning with Complementary Labels}

\appendix
\section{Proofs}
\subsection{Proof of Proposition \ref{p1}}
\begin{proof}
Let $\bm{\eta}$ and $\bm{\overline{\eta}}$ denote the conditional distribution
$\mathbb{P}(Y\mid X)$ and $\mathbb{P}(\overline{Y}\mid X)$ respectively, where
$\bm{\eta}_k(x)=\mathbb{P}(Y=k\mid x)$ and
$\bm{\overline{\eta}}_k(x)=\mathbb{P}(\overline{Y}=k\mid x)$.
Since $\bar{y}$ only depends on $y$, we have $\bar{\bm{\eta}}(x)=T^\top\bm{\eta}(x)$.
The unbiased risk estimator can be derived as follows:
\begin{align*}
    R(\bm{g};\ell)
    &= \mathbb{E}_{(x,y)\sim D} [\ell(y, \bm{g}(x))]
    = \mathbb{E}_{X} \mathbb{E}_{Y\sim\bm{\eta}(X)} [\ell(Y, \bm{g}(X))] \\
    &= \mathbb{E}_{X} [\bm{\eta}(X)^\top \ell(\bm{g}(X))]
    = \mathbb{E}_{X} [\bm{\overline{\eta}}(X)^\top (T^{-1}) \ell(\bm{g}(X))] \\
    &= \mathbb{E}_{(x, \overline{y})\sim \overline{D}} [e_{\overline{y}}^\top (T^{-1}) \ell(\bm{g}(x))]
\end{align*}
\end{proof}

\subsection{Proof of Proposition \ref{p2}}
\begin{proof}
Given the following two properties of ${\ell}_{01}$:
\begin{align*}
    \sum_{i=1}^{K}\ell_{01}(i, \bm{g}(x))=K-1\quad \text{ and} \\
    \ell_{01}(\overline{y}, \bm{g}(x))+\overline{\ell}_{01}(\overline{y}, \bm{g}(x))=1
\end{align*}
An unbiased risk estimator of classification error can be obtained by:
\begin{align*}
    R(\bm{g};\ell_{01}) 
    &= \mathbb{E}_{(x,\overline{y})\sim\overline{D}}
    \bigg[
    -(K-1)\ell_{01}(\overline{y},\bm{g}(x)) + \sum_{j=1}^{K}\ell_{01}(j,\bm{g}(x))
    \bigg] \\
    &= \mathbb{E}_{(x,\overline{y})\sim\overline{D}}
    \bigg[
    (K-1)(1-\ell_{01}(\overline{y},\bm{g}(x))) \rrbracket
    \bigg] \\
    &= (K-1) \mathbb{E}_{(x,\overline{y})\sim\overline{D}}
    \bigg[
    \overline{\ell}_{01}(\overline{y}, \bm{g}(x))
    \bigg] = (K-1)\overline{R}(\bm{g};\overline{\ell}_{01})
\end{align*}
\end{proof}

\subsection{Proof of Proposition \ref{p3}}
\begin{proof} The proposition can be derived by using the linearity of the gradient operator:
\begin{align*}
    \mathbb{E}_{\overline{y}\mid y}
    \big[
    \nabla_{\theta}\overline{\ell}(\overline{y}, \bm{g}(x))
    \big]
    &= \nabla_{\theta} \mathbb{E}_{\overline{y}\mid y}
    \big[
    \overline{\ell}(\overline{y}, \bm{g}(x))
    \big] \\
    &= \nabla_{\theta}
    \bigg[
    \frac{1}{K-1}\sum_{y'\neq y}
    \big[
    -(K-1)\ell(y',\bm{g}(x)) + \sum_{j=1}^{K}\ell(j,\bm{g}(x))
    \big]
    \bigg] \\
    &= \nabla_{\theta}
    \bigg[
    -\sum_{y'\neq y}\ell(y',\bm{g}(x))+
    \sum_{j=1}^{K}\ell(j,\bm{g}(x))
    \bigg] = \nabla_{\theta} \ell(y, \bm{g}(x))
\end{align*}
\end{proof}

\end{document}